\documentclass{svproc}

\usepackage{etoolbox}
\usepackage{hyperref}
\usepackage{cite}
\usepackage{pdfpages}
\usepackage[numbers,sort&compress]{natbib}
\usepackage{amsmath}
\usepackage{algorithm}
\usepackage{algpseudocode}
\usepackage{url}
\usepackage{graphicx}
\usepackage{tabularx}
\usepackage{float}
\usepackage{caption}
\usepackage{soul}
\usepackage{xcolor}
\sethlcolor{yellow} 
\usepackage{fancyhdr}
\usepackage{hyperref}

\hypersetup{
    colorlinks=true,
    linkcolor=blue,
    citecolor=blue,
    urlcolor=blue
}

\patchcmd{\thebibliography}{\newpage}{}{}{}

\begin{document}
\mainmatter         
\title{Extending XReason: Formal Explanations for Adversarial Detection}
\titlerunning{}  
%
\author{Amira Jemaa \and Adnan Rashid \and
Sofiène Tahar}
%
%
%
\institute{Department of Electrical and Computer Engineering \\
Concordia University, Montreal, QC, Canada,
\email{\texttt{} \\ \texttt{\{a\_jem, rashid, tahar\}@ece.concordia.ca}
}}

\maketitle 

\begin{abstract}
Explainable Artificial Intelligence (XAI) plays an important role in improving the transparency and reliability of complex machine learning models, especially in critical domains such as cybersecurity. Despite the prevalence of heuristic interpretation methods such as SHAP and LIME, these techniques often lack formal guarantees and may produce inconsistent local explanations. To fulfill this need, few tools have emerged that use formal methods to provide formal explanations. Among these, XReason uses a SAT solver to generate formal instance-level explanation for XGBoost models. In this paper, we extend the XReason tool to support LightGBM models as well as class-level explanations. Additionally, we implement a mechanism to generate and detect adversarial examples in XReason. We evaluate the efficiency and accuracy of our approach on the CICIDS-2017 dataset, a widely used benchmark for detecting network attacks.
\end{abstract}
\section{Introduction}
Artificial Intelligence (AI) has revolutionized various industries by automating complex tasks and enabling data-driven decision-making. However, as AI systems become more sophisticated, concerns about their transparency and trust have emerged, leading to the rise of Explainable Artificial Intelligence (XAI) \cite{xai1}. XAI allows the decision-making processes of AI models to be comprehended by human users which is a crucial requirement for raising trust in high-stakes domains such as healthcare, finance, and network security. Moreover, AI models remain vulnerable to adversarial attacks \cite{adversarial1}, where malicious inputs can manipulate outcomes. XAI plays a pivotal role in mitigating these risks by providing insights into model behavior, aiding in the detection of vulnerabilities, and strengthening the robustness and security of AI applications.

Well-known XAI tools, such as SHAP \cite{shap}, LIME \cite{lime}, are model-agnostic, meaning that they can be applied to any machine learning model without requiring access to the internal structure of the model. These tools work by treating the model as a black box and focusing on explaining the relationship between inputs and outputs. While model-agnostic methods are useful for a variety of models, they have significant limitations. In fact, these methods are often heuristic, meaning they can generate different explanations for the same prediction. Furthermore, they are not always correct, which can result in counterexamples, where the explanations do not hold.

On the other hand, formal methods in XAI have emerged as a robust alternative for generating precise and reliable explanations. Formal methods translate complex machine learning models into mathematical logic representations \cite{JoaoMarques-Silva}. Subsequently, formal methods ensure the soundness and completeness of explanations by identifying the minimal set of features responsible for the model’s decision. This approach strengthens the model’s defenses against adversarial attacks by accurately identifying exploitable features in its decision-making process. Moreover, the generation of adversarial examples, i.e., small perturbations in input data designed to mislead a model, highlights the importance of understanding the feature-level behavior. In this regard, the identification of the features influencing decisions of the model allows the generation of more targeted and effective adversarial samples. XAI is essential in this context as it reveals the feature importance, enabling precise manipulation of only those features that drive the model’s prediction. However, this strategy is only reliable when the truly relevant features are accurately identified. By rigorously analyzing the decision-making process, formal methods ensure that perturbations are applied to the correct features, improving both the generation and detection of adversarial examples.

A few formal XAI tools, such as XReason \cite{xreasonwebsite}, Silas \cite{silas} and PyXAI \cite{pyxai}, have been recently introduced to provide formal explanations. For instance, XReason uses XGBoost models \cite{chen2016xgboost} and SAT solvers \cite{sat} to provide detailed, instance-level explanations, while Silas focuses on Random Forest models \cite{breiman2001random} and employs SMT solvers \cite{de2008z3}. PyXAI, on the other hand, focuses on models such as RandomForest Classifier, XGBoost Classifier, XGBoost Regressor, and LGBM Regressor with the use of SAT solvers. After considering these options, we selected XReason due to its open-source nature and flexibility, allowing for customization and extensibility. While XReason is effective for XGBoost models, it lacks support for other more efficient machine learning models, such as LightGBM \cite{lightgbm}, which is a gradient boosting framework that is widely recognized for its efficiency and scalability, especially when dealing with large datasets and requiring faster training times. Additionally, XReason only provides instance-level explanations, which focus on single predictions without offering a broader understanding of model behavior across classes. To overcome these limitations, we propose in this paper to extend XReason by integrating support for LightGBM and introduce class-level explanations. These explanations identify the most important features for each class, providing a more comprehensive view of the model's decision-making process. Moreover, since XReason does not address adversarial robustness, we also propose to incorporate the ability to both generate and detect adversarial examples.

The rest of the paper is organized as follows: In Section \ref{sec:section2}, we explain the methods used in our framework for extending XReason. Section \ref{sec:section3} describes our approach for the propositional encoding of LightGBM. In Section \ref{sec:section4}, we discuss class-level explanations. In Section \ref{sec:section5}, we present our methods for generating and detecting adversarial examples. Finally, in Section \ref{sec:section6}, we present our experiments using the CICIDS-2017 dataset \cite{cicid}, where we evaluate the robustness and correctness of our formal explanation method in generating and detecting adversarial examples.
\section{Proposed Framework}
\label{sec:section2}
\begin{figure}[H]
    \centering
\includegraphics[width=\textwidth]{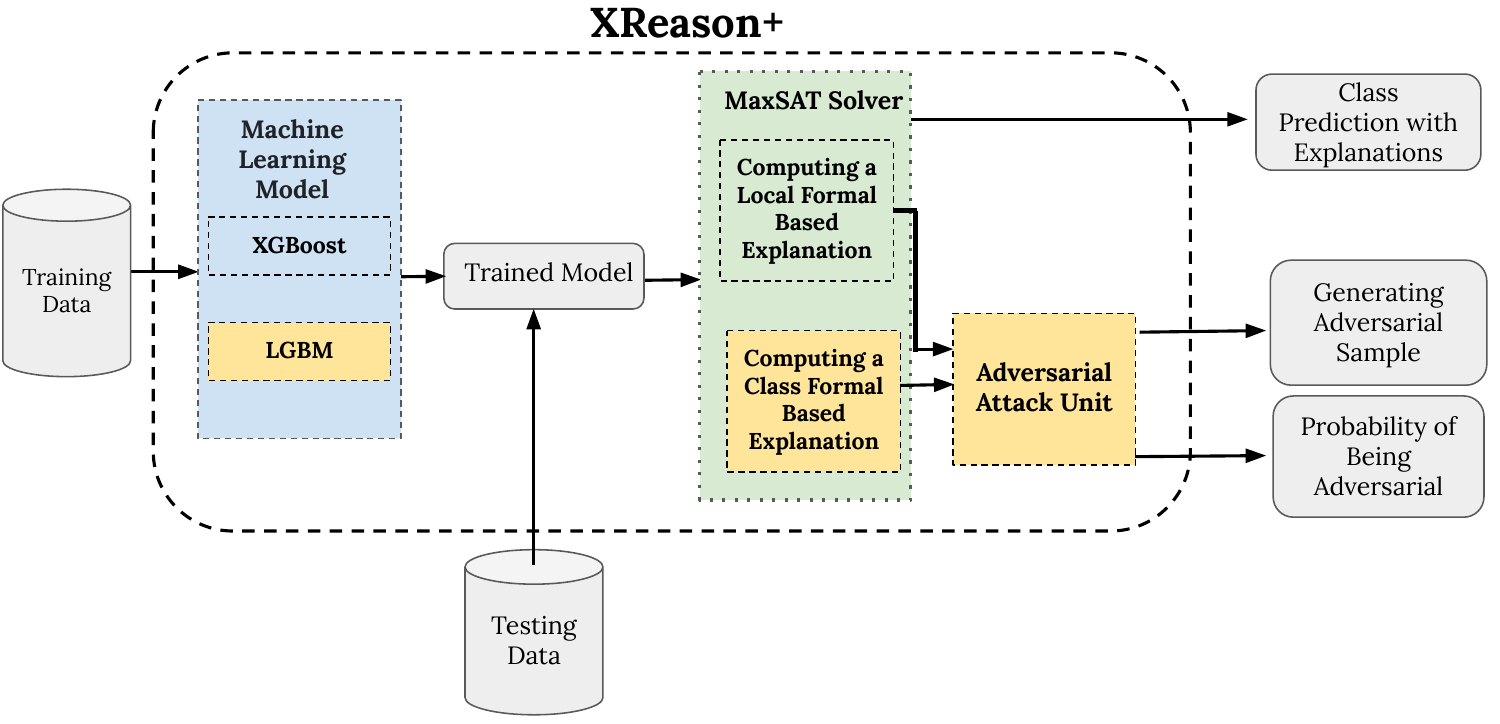}
    \caption{XReason+ Tool.}
    \label{fig:diagram}
\end{figure}

The proposed methodology builds on the existing tool XReason\cite{xreasonwebsite}, which addresses classification problems by providing abductive formal explanations (AXps) \cite{abductive} for XGBoost models using a SAT solver. XReason explains \textit{why} an XGBoost model makes a particular prediction for a given sample by identifying a minimal subset of features responsible for the decision. Figure \ref{fig:diagram} presents our proposed extensions to XReason, called XReason+ \footnote{\href{https://hvg.ece.concordia.ca/projects/fvai/pr2/}{https://hvg.ece.concordia.ca/projects/fvai/pr2/}}. Our framework enhances XReason's capabilities in several key areas: \begin{itemize}
    \item \textbf{Model Support:} We expand XReason to support LightGBM in addition to XGBoost, enabling formal reasoning and explanations across multiple machine learning models.
    \item \textbf{Class-Level Explanations:} We introduce class-level explanations, which create intervals for the most important features defining each class. This provides a broader view of model behavior compared to instance-based explanations, offering insights into how features contribute to predictions for each class.
    \item \textbf{Adversarial Sample Handling:} Using formal explanations, we implement an adversarial attack mechanism that can both generate and detect adversarial samples. Detection is based on calculating the probability of a sample being adversarial by analyzing changes in explanations in response to small input perturbations.
\end{itemize}
As depicted in Figure~\ref{fig:diagram}, the XReason+ process begins by training either XGBoost or LightGBM (LGBM) models on the provided training data. Once the model is trained, test data is processed to compute both instance-level and class-level formal explanations using a MaxSAT \cite{ignatiev2019rc2} solver. These explanations are then used to produce class predictions for the test data, accompanied by formal explanations. Moreover these explanations are used as input for the adversarial attack unit, which can either generate adversarial samples from the test data or detect the probability of the test data being adversarial by analyzing explanation changes in response to small perturbations.

\vspace{5 pt}

In the next section, we describe the formal encoding of LightGBM models in XReason. This encoding captures the exact paths leading to predictions, enabling formal analysis.
\section{Propositional Encoding of LightGBM}
\label{sec:section3}
LightGBM is a widely used gradient boosting framework designed for efficiency
and scalability. Its leaf-wise growth strategy differs from the level-wise approach
in traditional tree-based models, prioritizing leaves with the highest reduction in loss, resulting in deeper, more complex trees. While this design improves training time and accuracy, it complicates interpretability due to unbalanced tree structures.

In this section, we propose a formal encoding of LightGBM to produce precise explanations capturing the model’s decision-making process. Each decision tree in LightGBM can be represented as a series of logical constraints that encode the conditions (splits) at each node. For instance, a split condition like $f_1 > 0.5$ is encoded as a Boolean variable, where $f_1 = 1$ if the condition holds and $f_1 = 0$ otherwise. Each path through the tree represents a conjunction of these Boolean variables, and each leaf node contains the prediction. Formally, encoding a tree is a logical formula representing the conjunction of feature conditions leading to a particular leaf.

Algorithm~\ref{alg:encoding} outlines the steps to encode a decision tree into a logical representation in XReason. The algorithm starts by receiving a decision tree as input, which includes its features, thresholds (the values used to split data), and branches. The first step is to collect all the thresholds used in the tree for each feature. Then, for each of these thresholds, the algorithm assigns logical variables that represent the decision points in the tree where the data is split. Once the thresholds and logical variables are assigned, the algorithm moves to the branches of the tree. For each branch, it creates a logical path, which describes how the features and thresholds lead to a specific decision.
\vspace{-14 pt}
\begin{algorithm}[H]
\caption{Propositional Encoding of Decision Tree}
\label{alg:encoding}
\begin{algorithmic}[1]
    \State \textbf{Input:} Decision tree with nodes, features $f_i$, thresholds $t_i$, and branches $b_i$.
    \State \textbf{Output:} Encoded paths $P$ with logical constraints.
    \State Initialize $Thresholds \leftarrow \emptyset$, $Lvars \leftarrow \emptyset$, $Paths \leftarrow \emptyset$
    \For{each feature $f_i$}
        \State Extract thresholds $t_i$ and add to $Thresholds(f_i)$
        \State Assign logical variables $L_i$ for splits and add to $Lvars(f_i)$
    \EndFor
    \For{each branch $b_i$}
        \State Traverse and extract constraints, form path $p_i$ from $Lvars$, and add to $Paths$
    \EndFor
    \For{each path $p_i$ in $Paths$}
        \State Expand $p_i$ into logical constraints $[L_1, L_2, ..., L_n, 0]$ and add to final set
    \EndFor
    \For{each new path from further splits}
        \State Repeat the previous steps for new paths
    \EndFor
    \State Ensure paths respect tree order and encode feature domains with logical variables
    \State \textbf{Return:} Encoded paths $P$
\end{algorithmic}
\end{algorithm}

  These paths are then expanded into a set of logical rules or constraints that represent how the decision-making process works in the tree. If the tree splits further and creates new paths, the algorithm applies the same process to these additional paths. In the final steps, the algorithm ensures that the order of decisions in each path matches the structure of the tree and that the feature values are encoded correctly.
\section{Class-Level Explanations}
\label{sec:section4}
The Class-Level Explanation constructs explanations for each class in the model by aggregating important features from training instances that belong to a specific class. This process helps identify the common characteristics that define each class and can be used to understand the behavior of the model at the class level.

Algorithm~\ref{alg:class_level_explanation} details the procedure for building the class-level explanations in XReason. The algorithm begins by iterating over all classes in the trained model $M$. For each class $c$, it creates an empty set $\mathcal{E}_c$ to store the important features. It then analyzes each instance $x_i$ in the training dataset $D_{\text{train}}$, where the model predicts the class $c$ (i.e., $M(x_i) = c$). For each instance, a formal explanation method is used to extract the important features $\mathcal{F}_i$. These latter highlight the key input features that contributed to the model’s decision for that particular instance. The important features of each instance $x_i$ are added to the class-level explanation set $\mathcal{E}_c$, which accumulates the significant features for the entire class.

\vspace{- 1 em}
\begin{algorithm}[H] 
\caption{Class-Level Explanation Building}
\label{alg:class_level_explanation}
\begin{algorithmic}[1]
\Require Trained model $M$, training dataset $D_{\text{train}}$
\Ensure Class-level explanations $\mathcal{E}_c$ for each class $c$
\Procedure{BuildClassLevelExplanations}{}
    \For{each class $c$ in model $M$}
        \State Initialize $\mathcal{E}_c$ as an empty set
        \For{each instance $x_i$ in $D_{\text{train}}$ where $M(x_i) = c$}
            \State Extract important features $\mathcal{F}_i$ from formal explanation of $x_i$
            \State $\mathcal{E}_c \gets \mathcal{E}_c \cup \mathcal{F}_i$
        \EndFor
        \For{each important feature $f$ in $\mathcal{E}_c$}
            \State Collect all values $V_f^c$ of feature $f$ in $\mathcal{E}_c$
            \State Determine interval $[a_f^c, b_f^c]$ using clustering on $V_f^c$
        \EndFor
    \EndFor
\EndProcedure
\end{algorithmic}
\end{algorithm}
\vspace{- 1em}
Once the important features for all instances belonging to the class $c$ have been collected, the algorithm identifies the key ranges of values for each feature. For each important feature $f$ in $\mathcal{E}_c$, it collects all values $V_f^c$ observed across the instances. Using a clustering technique, the algorithm determines an interval $[a_f^c, b_f^c]$, which represents the typical range of feature values for the class $c$.
These intervals concisely represent the most relevant features for each class, highlighting how the model distinguishes between classes based on specific ranges. This information is valuable for model interpretation and downstream tasks, such as adversarial sample generation, where these feature intervals can help manipulate input data to mislead the model.
In next section, we propose an approach to generate and detect such adversarial attacks.
\section{Adversarial Example Detection Method}
\label{sec:section5}
Algorithm~\ref{alg:adversarial_detection} outlines the process of detecting adversarial examples by comparing the explanations of an input sample with the class-level explanations derived from the training data. The detection relies on two checks: verifying the importance of the features and ensuring that their values fall within expected intervals for the predicted class.

The process begins by predicting the label of the input sample, $x_{\text{input}}$, using the trained model $M$. Once the label is predicted, the important features and their values are extracted from the formal explanation of $x_{\text{input}}$. For the predicted class, class-level explanations $\mathcal{E}_{y_{\text{input}}}$ are retrieved, which contain the important features and their typical intervals.

The algorithm compares the input sample’s features with class-level explanations by performing two checks for each important feature: first, it verifies if the feature is important for the predicted class by checking its presence in the class-level explanation; if the feature is identified as important, the algorithm then checks whether its value falls within the interval defined in the class-level explanation for that class.
If a feature is either not present in the class-level explanation or its value falls outside the defined interval, it is considered a discrepancy. The number of such discrepancies is accumulated, and the adversarial likelihood score $s_{\text{adv}}$ is computed as the ratio of discrepancies to the total number of important features in the input sample.
\vspace{-1 em}
\begin{algorithm}[H]
\caption{Adversarial Sample Detection Using Class-Level Explanations}
\label{alg:adversarial_detection}
\begin{algorithmic}[1]
\Require Trained model $M$, class-level explanations $\{\mathcal{E}_c\}$, input sample $x_{\text{input}}$
\Ensure Adversarial likelihood score $s_{\text{adv}}$
\State $y_{\text{input}} \gets M(x_{\text{input}})$ \Comment{Predict label of input sample}
\State Extract important features $\mathcal{F}_{\text{input}}$ and their values $V_{\text{input}}$ from explanation of $x_{\text{input}}$
\State Retrieve class-level explanations $\mathcal{E}_{y_{\text{input}}}$
\State Initialize discrepancy count $d \gets 0$
\For{each feature $f$ in $\mathcal{F}_{\text{input}}$}
    \If{$f$ is not in $\mathcal{E}_{y_{\text{input}}}$}
        \State $d \gets d + 1$ \Comment{Feature not expected to be important}
    \Else
        \State Retrieve interval $[a_f^{y_{\text{input}}}, b_f^{y_{\text{input}}}]$
        \If{$V_{\text{input}}(f) \notin [a_f^{y_{\text{input}}}, b_f^{y_{\text{input}}}]$}
            \State $d \gets d + 1$ \Comment{Value outside expected interval}
        \EndIf
    \EndIf
\EndFor
\State Compute adversarial likelihood score $s_{\text{adv}} \gets \dfrac{d}{|\mathcal{F}_{\text{input}}|}$
\State \Return $s_{\text{adv}}$
\end{algorithmic}
\end{algorithm}
\vspace{- 1 em}

\section{Case Study: CICIDS-2017 Dataset}
\label{sec:section6}
As a case study for our proposed XReason+ tool, we utilize a customized version of the Canadian Institute for Cybersecurity’s Intrusion Detection System 2017 (CICIDS-2017) dataset \cite{cicid}, which is widely used for network security research. The original dataset simulates both normal network traffic and various types of network attacks, making it highly representative of real-world conditions. It contains over 3 million records, 80 network features and 14 attack types, with an imbalanced class distribution. This dataset is particularly suitable for classification tasks, as it includes a labeled target variable indicating whether each instance represents normal traffic or an attack.
We adopt the modified version of CICIDS-2017 proposed by Li et al. \cite{li_cicid}, which reduces the feature set to 19 essential attributes, selected for their relevance to network traffic patterns.
\subsection{Preprocessing and Model Performance}
To ensure the quality and relevance of the data, we applied additional preprocessing steps by first removing any duplicate records in the dataset, followed by feature selection using the Autospearman method \cite{autospearman} to automatically eliminate highly correlated features, resulting in a reduced set of 18 critical features. Finally, we split the dataset into 70\% for training (20,655 samples) and 30\% for testing (8,853 samples).
We trained a LightGBM model, following the recommendation of Li et al.\cite{li_cicid}, and evaluated the model’s performance on the testing set using the key metrics: Accuracy, Precision, Recall, F1 Score, and Area Under the ROC Curve (AUC) \cite{roc}. The respective values obtained range from 90\% to 93\%.
\subsection{Robustness and Correctness of Formal Explanations}
Many previous studies have used SHAP and LIME to explain models trained on CICIDS-2017 data (e.g., \cite{cicid_shap, cicid_lime}). While these methods provide valuable insights into feature
importance, they are prone to instability and lack formal guarantees. In contrast, our formal explanation method provides consistent and provably correct explanations.

\subsubsection{Robustness of Explanations}
To assess the robustness of these methods, we applied SHAP, LIME, and our formal approach on the same instances twice to evaluate the consistency of feature values and rankings across both runs. SHAP demonstrated 100\% consistency, producing identical feature importance scores and rankings on repeated runs. In contrast, LIME showed 0\% consistency, with feature rankings and importance scores varying each time, indicating high variability and limited reliability. Our formal explanation approach, XReason+, is fully deterministic, consistently identifying the same features and ranks for each instance across runs, thereby demonstrating great robustness.
\subsubsection{Correctness of Explanations} The formal explanation method provides 100\% correct explanations by identifying minimal feature sets responsible for predictions. We evaluate how closely SHAP and LIME rankings align with those from our formal method. In the formal explanation ranking, features in the explanation are assigned rank 1, with all other features following in the next rank. SHAP and LIME ranks were compared to the formal method using Spearman’s Rank Correlation \cite{spearman}, Kendall’s Tau \cite{kendall}, and Rank-Biased Overlap (RBO) \cite{rbo}. For Spearman, SHAP values ranged from -0.28 to 0.59 (avg 0.14), and LIME from -0.70 to 0.72 (avg 0.18). Kendall’s Tau showed SHAP from -0.24 to 0.49 (avg 0.11) and LIME from -0.59 to 0.60 (avg 0.15). RBO averaged 0.39 for SHAP and 0.37 for LIME. In conclusion, our formal method proves more robust and correct than SHAP and LIME. Given this strong foundation of reliability and correctness, we can use the key features identified through our method to craft and detect adversarial samples that exploit the model's vulnerabilities.
\subsection{Adversarial Sample Generation and Detection }
To evaluate the robustness of our approach, we applied the adversarial sample generation method across the entire test set, consisting of 8,853 samples. Out of these, 2,821 samples (31.86\%) successfully fooled the model, highlighting its vulnerability to adversarial attacks. The average Euclidean distance between the original and adversarial samples was 1.67, indicating that only small perturbations were needed to manipulate the model's predictions.
Further analysis showed a significant difference in the impact of adversarial attacks based on the original class. Out of the 3,058 samples that were originally predicted as attacks (class 1), 2,044 (66.84\%) were misclassified as normal traffic (class 0) after the perturbation. Moreover, 777 out of the 5,795 samples originally predicted as normal (class 0) were flipped to attacks (class 1), which represents 13.41\%.
Using our formal explanation-based detection method, we were able to identify 1,731 out of the 2,821 adversarial examples (those that changed the model’s prediction) as likely adversarial, achieving a detection rate of 61.36\%. This demonstrates the effectiveness of the detection mechanism in identifying samples that exploit model vulnerabilities, particularly in the case of misclassified attack samples.
\section{Conclusion}
In this paper, we proposed a framework extending the XReason tool by adding formal explanations for LightGBM models, class-level explanations, and a method to generate and detect adversarial samples. Our approach provides deterministic and consistent explanations, addressing the limitations of heuristic methods like SHAP and LIME. We applied our  XReason+ tool to the CICIDS-2017 dataset, demonstrating its effectiveness in generating adversarial samples and improving robustness.
In the future, we plan to extend the  developed framework to support additional machine learning models and apply it to other large-scale datasets. We also aim to enhance the adversarial detection mechanism.

%
%
\bibliographystyle{splncs03}
\bibliography{271}

\end{document}